\documentclass{article}
\usepackage{ijcai25}

\usepackage{times}
\usepackage{soul}
\usepackage{url}
\usepackage[hidelinks]{hyperref}
\usepackage[utf8]{inputenc}
\usepackage[small]{caption}
\usepackage{graphicx}
\usepackage{amsmath}
\usepackage{amsthm}
\usepackage{booktabs}
\usepackage{algorithm}
\usepackage{algorithmic}
\usepackage[switch]{lineno}

\title{Chat-of-Thought: Collaborative Multi-Agent System for Generating Domain Specific Information}

\author{
Christodoulos Constantinides$^1$
\and
Shuxin Lin$^2$
\and
Nianjun Zhou$^2$
\and
Dhaval Patel$^{2}$
\affiliations
$^1$IBM\\
$^2$IBM Research\\
\emails
\{christodoulos.constantinides@, shuxin.lin@, jzhou@us., pateldha@us.\}ibm.com,
}

\begin{document}

\maketitle

\begin{abstract}
This paper presents a novel multi-agent system called Chat-of-Thought, designed to facilitate the generation of Failure Modes and Effects Analysis (FMEA) documents for industrial assets. Chat-of-Thought employs multiple collaborative Large Language Model (LLM)-based agents with specific roles, leveraging advanced AI techniques and dynamic task routing to optimize the generation and validation of FMEA tables. A key innovation in this system is the introduction of a \textit{Chat of Thought}, where dynamic, multi-persona-driven discussions enable iterative refinement of content. This research explores the application domain of industrial equipment monitoring, highlights key challenges, and demonstrates the potential of Chat-of-Thought in addressing these challenges through interactive, template-driven workflows and context-aware agent collaboration.
\end{abstract}

\section{Introduction}

In modern data-driven industries, teams often rely on diverse expertise to extract actionable insights and optimize operations. Collaborative efforts between subject matter experts, data scientists, and engineers play a critical role in analyzing complex systems, identifying risks, and improving decision-making. However, such processes are resource-intensive and require significant coordination. To address these challenges, we propose a novel framework, Chat-of-Thought, which leverages the role-playing capabilities of Large Language Models (LLMs) \cite{achiam2023gpt}, \cite{touvron2023llama}, \cite{jiang2023mistral} to simulate collaborative, multi-agent environments \cite{park2023generative} for domain-specific knowledge generation.

The Chat-of-Thought system enables the creation of virtual personas, each representing a specific area of expertise. These personas operate collaboratively to generate hypothetical documents and insights, without requiring human intervention. By integrating inputs from diverse information sources—such as Failure Modes and Effects Analysis (FMEA), Key Performance Indicators (KPIs), diagnostic data, YAML configurations, synthetic datasets, and modeling code—this framework facilitates the exploration of complex scenarios in an automated and scalable manner.

\begin{figure}[t!]
    \centering
    \includegraphics[width=0.5\linewidth]{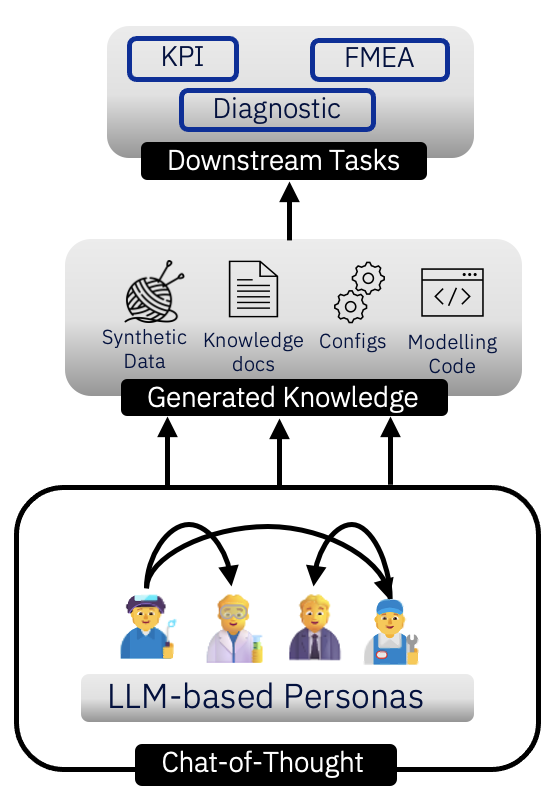}
    \caption{Overview of Chat-of-Thought.}
    \label{fig:enter-label}
\end{figure}

A key application of this framework is the automation of FMEA, a systematic reliability engineering method used to identify and mitigate potential failure modes in industrial systems. The manual creation of FMEA documents is labor-intensive and demands multidisciplinary expertise, making it an ideal candidate for automation. Our system, Chat-of-Thought, employs a multi-agent architecture with specialized roles to streamline FMEA generation, ensuring accuracy and scalability while significantly reducing the effort required.

By combining the collaborative potential of simulated personas with diverse data inputs, the Chat-of-Thought system demonstrates a transformative approach to process optimization, risk identification, and decision-making. This work aims to illustrate the potential of LLMs to revolutionize traditional workflows, enabling industries to harness artificial intelligence for enhanced reliability and operational efficiency.

\begin{figure*}[th!]  
    \centering  
    \includegraphics[width=0.95\linewidth]{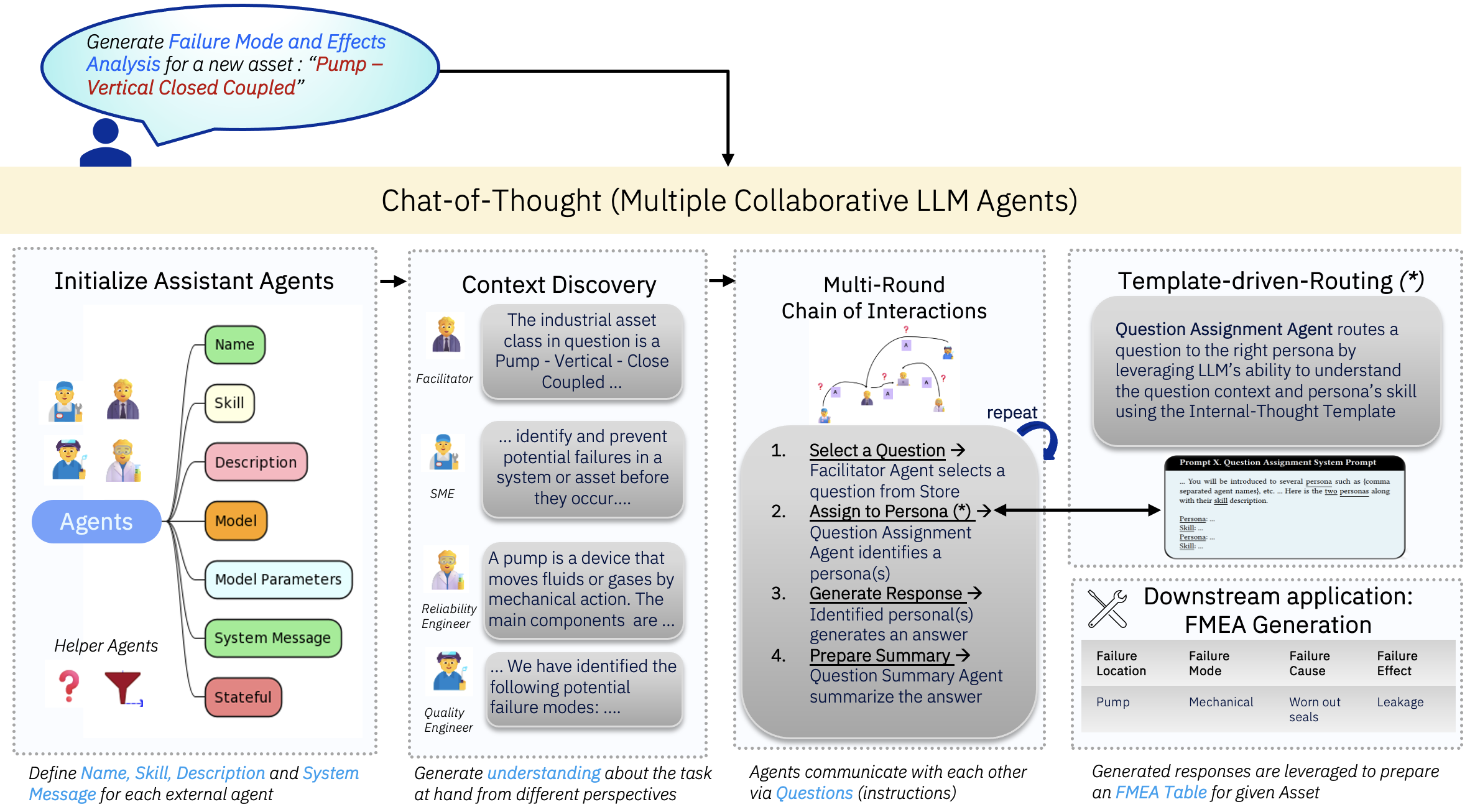}   
    \caption{Overview of Chat-of-Thought System Architecture}  
    \label{fig:system-architecture}  
\end{figure*}  

\subsection{Application Domain and Problem Scenario}  
The system is designed for industries managing complex and high-value assets, including pumps, boilers, chillers, and similar equipment. These assets are inherently prone to failures caused by mechanical, electrical, and operational issues, often resulting in significant downtime and costs. The manual creation of Failure Modes and Effects Analysis (FMEA) documents, a critical task in reliability engineering, suffers from limitations such as lack of scalability, inconsistencies, and delays in validation—particularly for out-of-scope (OOS) scenarios involving novel or less-common assets. The primary challenges in this domain include:  
\begin{itemize}  
    \item The requirement for highly specialized domain-specific expertise.  
    \item Accurate contextual understanding of diverse asset classes and their associated failure mechanisms.  
    \item Streamlined and efficient validation of FMEA data to minimize the workload of Subject Matter Experts (SMEs).  
\end{itemize}  

\section{System Architecture and Key Components}  
The proposed \textbf{Chat-of-Thought} system employs a collaborative multi-agent framework powered by Large Language Models (LLMs). Each agent within the system is designed with predefined roles, skills, and contextual system messages, enabling them to simulate domain-specific expertise. The architecture is structured into the following key stages:  

\subsection{Initialization of Assistant Agents}  
Agents are initialized with predefined attributes that simulate specific roles essential for FMEA generation. These roles include, but are not limited to:  
\begin{itemize}  
    \item \textbf{Facilitator:} Oversees the overall interaction and ensures task progression.
    \item \textbf{Reliability Engineer:} Focuses on identifying potential failure modes and their underlying mechanisms.  
    \item \textbf{Quality Engineer:} Reviews generated insights for accuracy and consistency.  
    \item \textbf{SME Validator:} Validates outputs to ensure they meet domain-specific standards. 
    \item \textbf{Summarizer:} Summarizes the document and answers into a concise and readable format.
\end{itemize}  
Each agent \( A_i \) is represented as a tuple \( (R_i, S_i, M_i) \), where \( R_i \) denotes the role, \( S_i \) represents the skill set, and \( M_i \) specifies the system messages guiding the agent’s behavior.  

\subsection{Context Discovery}  
The system begins by identifying the asset class (e.g., Pump - Vertical Close-Coupled) and its operational parameters. Relevant failure modes, their causes, and associated effects are dynamically extracted from domain-specific knowledge repositories and historical data. This stage ensures that the analysis is tailored to the specific context of the asset being examined.  

\begin{figure*}[h!]
    \centering
    \includegraphics[width=0.9\linewidth]{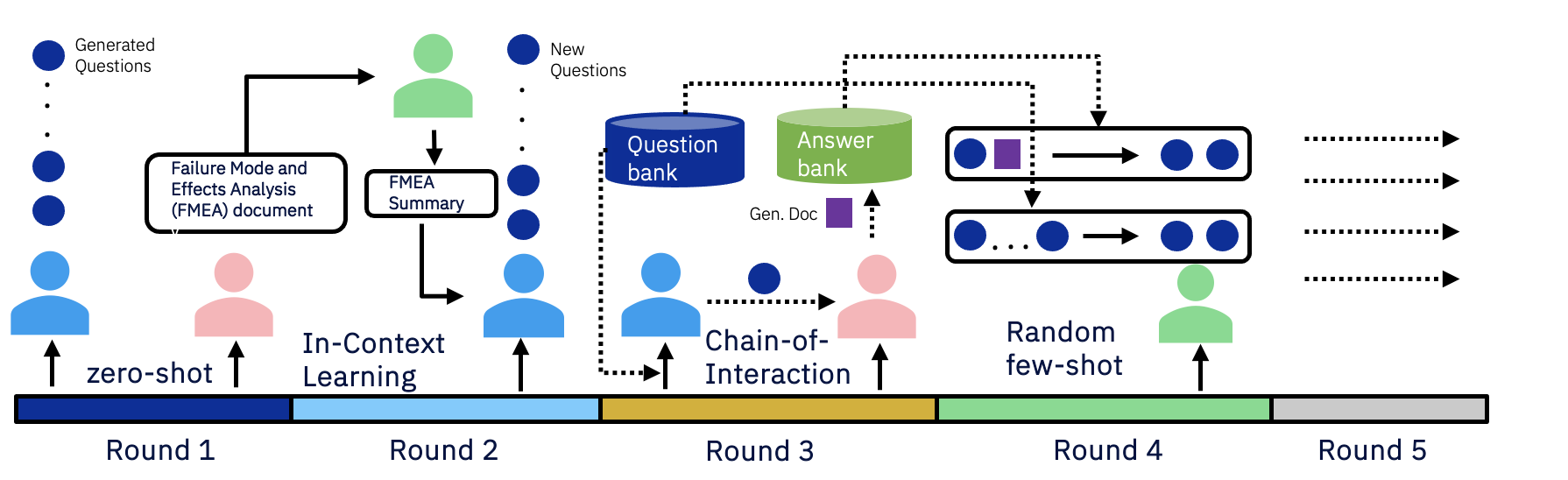}
    \caption{Round progression. Chat-of-Thought is a flexible framework where agents collaborate using various methods.}
    \label{fig:enter-label}
\end{figure*}

\subsection{Multi-Round Chain of Interactions}  
The core innovation of Chat-of-Thought lies in its \textit{Chat-of-Thought} mechanism, where agents engage in iterative, multi-persona-driven discussions to collaboratively refine outputs. This approach extends beyond the traditional \textit{Chain-of-Thought} framework by fostering dynamic, context-aware dialogue across specialized personas. The process consists of the following steps:  
\begin{enumerate}  
    \item \textbf{Question Selection:} The Facilitator Agent selects a question \( Q \) from a predefined question bank \( Q_B \).  
    \item \textbf{Dynamic Persona Assignment:} The Question Assignment Agent routes the selected question to the most suitable persona \( A_i \) based on contextual understanding \( C \) and skill alignment \( S_i \).  
    \item \textbf{Response Generation:} The assigned persona generates a response \( R \), incorporating its domain-specific expertise and LLM capabilities.  
    \item \textbf{Summary Aggregation:} The Question Summary Agent synthesizes responses into a cohesive and actionable summary for downstream processing.  
\end{enumerate}  

\subsection{Template-Driven Routing}  
To ensure high-quality outputs, the system employs a template-based routing mechanism. Each template \( T \) defines structured guidelines for routing questions to the appropriate personas. These templates are dynamically adjusted to maintain contextual relevance and optimize the quality of generated FMEA data. By leveraging this template-driven approach, Chat-of-Thought ensures consistency across interactions while adapting to the nuances of various scenarios.  

This architecture, with its emphasis on collaborative intelligence and role-specific expertise, empowers industries to automate labor-intensive FMEA tasks effectively while maintaining precision and scalability.

\subsection{Quality check}
At the end of each round there is a quality control to filter out useless questions using a pre-trained classifier, and remove duplicates questions/answers using a self-BLEU score \cite{papineni2002bleu} threshold. 

\section{AI Techniques and Innovations}  
Chat-of-Thought integrates advanced AI techniques to streamline FMEA generation and enhance performance in complex tasks. At its core, the system leverages the power of Large Language Models (LLMs) for domain-specific reasoning, natural language understanding, and generating high-quality outputs. Task assignments are guided by context-aware routing mechanisms, where internal-thought templates align agent skills with the query's specific requirements. Multi-agent collaboration ensures iterative refinement of outputs, while a stateful design allows the agents to manage multi-step and context-dependent processes efficiently.  

The system operates through a multi-phase process that combines failure model analysis, iterative learning, and question-answering frameworks. The initial phase (\textbf{Round 1}) evaluates zero-shot learning capabilities, testing the system’s baseline performance without task-specific examples. This is followed by in-context learning (\textbf{Round 2}), where scenarios derived from Failure Modes and Effects Analysis (FMEA) guide response generation and incorporate human feedback for refinement.  

Subsequent rounds build on this foundation (\textbf{Round 3}). Chain-of-interaction techniques introduce feedback loops between the question bank and answer bank, enabling iterative learning and continuous improvement. Random few-shot learning methods (\textbf{Round 4}) enhance the system’s ability to generalize knowledge from limited samples, making it adaptable to new or uncommon scenarios. These iterative stages culminate in a robust and optimized system capable of addressing complex, domain-specific challenges.  
\section{Result and Demonstration}  
The demonstration of Chat-of-Thought highlights its capabilities across various operational stages. Agents are initialized with distinct roles and skills to collaboratively generate FMEA content. The \textit{Chat-of-Thought} process is showcased, where iterative multi-agent discussions refine the outputs and ensure contextually appropriate responses. The system also includes an interface for SME review, enabling efficient validation of the generated FMEA tables.  

Chat-of-Thought demonstrates its capability to generate detailed and accurate FMEA tables for both standard and out-of-scope assets. Validation by SMEs confirms that the system reliably identifies failure modes, root causes, and potential effects. These outputs facilitate proactive maintenance strategies and enhance overall reliability planning within industrial applications.  

\section{Demo}  
In the supplementary materials we provide a video with the walk-through of the system.

\nocite{autoq}
\nocite{selfinstruct}
\nocite{wu2023autogen}
\nocite{sun2023principle}
\nocite{wiki:Cross-industry-standard-process-for-data-mining}
\nocite{10.1145/3580305.3599827}
\nocite{uptake}
\nocite{He2023}
\nocite{li2023open}
\nocite{li2015diversity}
\nocite{ICWSM18LearningQ}
\nocite{kitaev2019multilingual}
\nocite{gong-etal-2022-khanq}
\nocite{cao2021controllable}
\nocite{ibmgenerativeaisdk}
\nocite{2022.EDM-posters.85}
\nocite{NEURIPS20228bb0d291}
\nocite{sanseviero2023moe}


\newpage

\bibliographystyle{named}
\bibliography{ijcai25}

\end{document}